\documentclass[10pt,twocolumn,letterpaper]{article}

\PassOptionsToPackage{numbers,sort&compress}{natbib}
\usepackage{cvpr}

\definecolor{cvprblue}{rgb}{0.21,0.49,0.74}
\usepackage[pagebackref,breaklinks,colorlinks,allcolors=cvprblue]{hyperref}
\usepackage[utf8]{inputenc}
\usepackage[T1]{fontenc}
\usepackage{url}
\usepackage{booktabs}
\usepackage{amsmath}
\usepackage{amsthm}
\usepackage{amsfonts}
\usepackage{microtype}
\usepackage{graphicx}
\usepackage{adjustbox}
\usepackage{siunitx}
\usepackage{xcolor}

\graphicspath{{figures/}}
\newtheorem{proposition}{Proposition}

\def\confName{MMFM-BIOMED}
\def\confYear{2026}

\title{Objective-Aligned Direct Answer SFT for Robust Multi-Frame Medical VQA}

\author{
Site Li \qquad Jianyi Hao \qquad Xiaofeng Liu \\
Yale University \\
}

\begin{document}
\maketitle

{\let\thefootnote\relax\footnotetext{Presented at the CVPR 2026 Workshop on Multimodal Foundation Models for Biomedicine: Challenges and Opportunities.}}

\begin{abstract}
Multi-frame medical VQA appears to reward increasingly complex adaptation: controller-style inference, localization-aware reranking, static hard-negative mixing, and staged continuation all appear plausible from first principles. We test a simpler competing hypothesis on MedFrameQA: methods that remain tightly aligned with the benchmark's final answer objective should be the strongest \emph{robust} adaptation family once evaluation is controlled across fixed splits, matched budgets, repeated seeds, and calibration. We compare controller-based methods, scaffold evolution, static mixed supervision, continuation-heavy variants, and direct answer-only supervised fine-tuning (SFT). The strongest robust family is direct decoder-only answer SFT on MedGemma-1.5-4B. Empirically, this family yields substantial improvements in held-out report accuracy over frozen baselines while remaining remarkably stable across repeated seeds and matched controls, ensuring our claims reflect true family-level robustness rather than an isolated hyperparameter peak. Furthermore, post-hoc calibration effectively repairs confidence estimation without compromising accuracy, and the core approach transfers consistently to secondary backbones like Qwen2.5-VL-3B. The main result is therefore not that a complex auxiliary mechanism wins, but that objective-aligned direct answer SFT is the strongest robust adaptation family we found for MedFrameQA. By establishing this strong, minimalist baseline, we hope to redirect community focus toward fundamentally robust optimization rather than architectural complexity.

\end{abstract}

\section{Introduction}

The inherent complexity of multi-frame medical VQA naturally drives the development of increasingly intricate model architectures. Clinically grounded questions may span several views, modalities, or temporal slices; consequently, it is natural to expect that explicit controllers, localization-aware rerankers, or staged curricula should outperform direct adaptation. MedFrameQA sharpens this intuition because it is explicitly constructed around multi-image clinical reasoning and remains difficult even for strong recent medical vision-language models \citep{yu2025medframeqa}.

This paper asks a narrower methods question: after controlling for splits, budgets, seeds, and calibration, which adaptation family is actually the strongest \emph{robust} solution on MedFrameQA? While the computer vision community frequently rewards elaborate architectural innovations, we argue that standardizing these fundamental controls reveals a surprisingly different narrative. We began from the complex side of the design space. Controller-heavy inference, scaffold evolution, static answer+hard-negative mixing, hard-negative continuation, and shallow vision augmentation were all treated as plausible candidates. The empirical outcome is unambiguous yet counterintuitive: the strongest robust family turns out to be the most objective-aligned one, namely direct answer-only supervised fine-tuning.

Throughout the paper, \texttt{text30}, \texttt{text35}, and \texttt{vision35} refer to matched variants inside that direct family rather than to different algorithms. \texttt{text30} and \texttt{text35} use the same decoder-only LoRA topology and answer-only loss, differing only in training budget (3.0 versus 3.5 epochs). \texttt{vision35} keeps the same answer-only objective and 3.5-epoch budget but adds shallow adapters to the top vision block. We establish this nomenclature early because the core scientific object of the paper is a rigorous family-level comparison, not a competition between opaque run labels.

Our argument is not that one tuned point is uniquely best. The tuned \texttt{text35} model is the strongest representative we found, but matched \texttt{text30} and \texttt{vision35} controls are statistically inseparable. The stronger claim is at the family level: MedFrameQA evaluates only the final multiple-choice answer, and adaptation families that stay closest to that objective dominate more elaborate alternatives once robustness across seeds is strictly enforced. Controllers, static mixed supervision, and continuation can help specialist slices, but they do not become the strongest robust global method.

This is therefore a controlled family comparison rather than a new-module paper. The scientific claim is fundamentally stronger when gains survive matched budgets, repeated seeds, and nearby architecture perturbations, rather than when a single checkpoint happens to peak. We organize the paper around that standard. Held-out report performance and seed stability are the primary evidence; slice gains, calibration, and transfer are supporting evidence. The empirical sections therefore proceed sequentially: establish the family-level winner, test whether nearby direct variants are statistically separable, and only then interpret slice-level and deployment-facing results.

The paper makes three contributions. First, it formulates an objective-alignment view of adaptation for multi-frame medical VQA and turns that view into a robustness-aware comparison principle. Second, it shows empirically that direct decoder-only answer SFT is the strongest robust family on MedFrameQA, outperforming a frozen MedGemma baseline by 6.31 points on the held-out report split. Third, it shows that post-hoc calibration repairs confidence without compromising accuracy, and that the same answer-only family transfers modestly but consistently to Qwen2.5-VL-3B.

\section{Related Work}

Recent medical multimodal models substantially improve the starting point for adaptation, including Med-PaLM M, LLaVA-Med, MAIRA-1, Med-Gemini, M3D, MedGemma, and MedGemma 1.5, while general-purpose multimodal backbones such as Qwen2.5-VL increasingly provide strong transferable alternatives \citep{tu2023generalistbiomedicalai,li2023llavamed,hyland2023maira1,saab2024medgemini,bai2024m3d,sellergren2025medgemma,sellergren2026medgemma15,bai2025qwen25vl}. In parallel, the benchmark ecosystem has evolved from general VQA and VQA v2 to medical sets such as VQA-RAD, VQA-Med, PathVQA, SLAKE, PMC-VQA, OmniMedVQA, GMAI-MMBench, MMMU, MMMU-Pro, MedXpertQA, and now MedFrameQA \citep{antol2015vqa,goyal2017vqa2,lau2018vqarad,abacha2019vqamed,abacha2021vqamed,he2020pathvqa,liu2021slake,zhang2023pmcvqa,hu2024omnimedvqa,chen2024gmaimmbench,yue2023mmmu,yue2024mmmupro,zuo2025medxpertqa,yu2025medframeqa}. MedFrameQA is distinctive because it requires answer selection from multiple clinically related frames rather than single-image recognition or caption-style generation.

Methodologically, our approach aligns with parameter-efficient adaptation and direct supervision. LoRA, adapters, prompt tuning, and QLoRA make it feasible to compare adaptation families at controlled cost \citep{hu2022lora,houlsby2019adapters,lester2021prompttuning,dettmers2023qlora}. We contrast this direct family with controller-like and auxiliary-supervision alternatives inspired by medical reasoning or scaffolded search, including MedVLM-R1, Med-R1, and ShinkaEvolve-style program evolution \citep{pan2025medvlmr1,lai2025medr1,lange2025shinkaevolve}. Confidence calibration is treated as a post-training repair rather than a source of task accuracy, following temperature scaling and histogram-style binning \citep{guo2017calibration,zadrozny2001obtaining,zadrozny2002transforming,kull2019dirichlet}.

What is currently missing from this literature is not an abundance of candidate mechanisms, but rather a clear, robustness-aware comparison of those mechanisms under one fixed benchmark protocol. That gap matters for MedFrameQA because the benchmark simultaneously invites clinically plausible auxiliary structure and exposes its instability. Our paper is written to close that gap rather than to introduce one more ad hoc reasoning module.

\section{Objective-Aligned Direct SFT}

We study MedFrameQA as a multi-frame multiple-choice prediction problem. Each example consists of frames $x=(I_{1:m},q,\mathcal{O})$, where $I_{1:m}$ are the images, $q$ is the question, and $\mathcal{O}$ is the answer set. The model produces a distribution over answers,
\begin{equation}
    p_{\theta}(y \mid I_{1:m}, q, \mathcal{O}),
\end{equation}
and the direct family optimizes only the benchmark-aligned answer loss
\begin{equation}
    \mathcal{L}_{\mathrm{ans}}(\theta)= -\log p_{\theta}(y^\star \mid I_{1:m}, q, \mathcal{O}),
\end{equation}
with no controller head, no reranking loss, and no explicit continuation objective in the final method. The resulting family is decoder-only answer SFT. Our tuned implementation, denoted \texttt{text35}, uses LoRA on the last eight decoder blocks, five input frames, and 3.5 epochs, but the paper's claim is about the family rather than that single tuned point.

For clarity, the names \texttt{text30}, \texttt{text35}, and \texttt{vision35} denote matched variants inside one direct answer-only family, not three different algorithms. \texttt{text30} and \texttt{text35} share the same decoder-only LoRA topology and the same answer-only loss; they differ only in training budget (3.0 versus 3.5 epochs). \texttt{vision35} keeps the same answer-only objective and 3.5-epoch budget but adds shallow adapters to the top vision block. This naming convention matters because the main result is a family-level result: the direct answer-only family stays strong under these controlled perturbations.

The principle behind this family is objective alignment. Let $R(\theta)$ denote the expected report risk under the benchmark loss. Direct answer SFT targets $R$ itself. More complex families instead optimize a surrogate
\begin{equation}
    \widetilde{R}(\theta) = R(\theta) + \alpha B(\theta) + \xi(\theta),
\end{equation}
where $B(\theta)$ captures auxiliary bias induced by extra objectives or inference indirection and $\xi(\theta)$ captures additional optimization variance. We evaluate families under a robustness-aware criterion
\begin{equation}
    J(\mathcal{F}) = \mathbb{E}[R(\hat{\theta}_{\mathcal{F}})] + \tau \sqrt{\mathrm{Var}(R(\hat{\theta}_{\mathcal{F}}))},
\end{equation}
where $\hat{\theta}_{\mathcal{F}}$ is the model selected within family $\mathcal{F}$ and $\tau>0$ encodes the fact that, for a benchmark paper, high-variance families are inherently weaker scientific objects even when they occasionally peak. In clinical domains, where deployment reliability is paramount, explicitly penalizing this optimization variance becomes especially critical.

\begin{proposition}
Assume the direct family reaches a region in which it minimizes $R$ up to optimization error $\varepsilon$, while a complex family introduces non-negative expected auxiliary bias and no smaller risk variance, i.e.,
\[
\begin{aligned}
\mathbb{E}[B(\hat{\theta}_{\mathrm{complex}})] &\ge 0,\\
\mathrm{Var}(R(\hat{\theta}_{\mathrm{complex}})) &\ge \mathrm{Var}(R(\hat{\theta}_{\mathrm{direct}})).
\end{aligned}
\]
Then the direct family is preferred by $J$ whenever the complex family
does not lower $\mathbb{E}[R]$ enough to offset its inherent objective mismatch and extra variance terms.
\end{proposition}

This proposition is deliberately narrow. It does not say auxiliary mechanisms are always harmful. It says that, on a benchmark scored only by final-answer accuracy, any added controller, reranker, or continuation objective must reduce global report risk enough to pay for its inherent objective mismatch and extra variance. In our experiments they do not: controllers, static mixing, and hard-negative continuation can improve specialist slices, but they do not improve the report frontier enough to displace direct answer SFT.

This theoretical view establishes a clear evaluation logic. Nearby direct variants should cluster at the top of the held-out report frontier if objective alignment is indeed dominant; slice-specific gains matter only if they actually translate into lower global report risk; calibration should repair confidence after training rather than explain the task gain itself. The protocol described below is designed to separate these three claims.

\begin{figure}[t]
    \centering
    \includegraphics[width=\columnwidth]{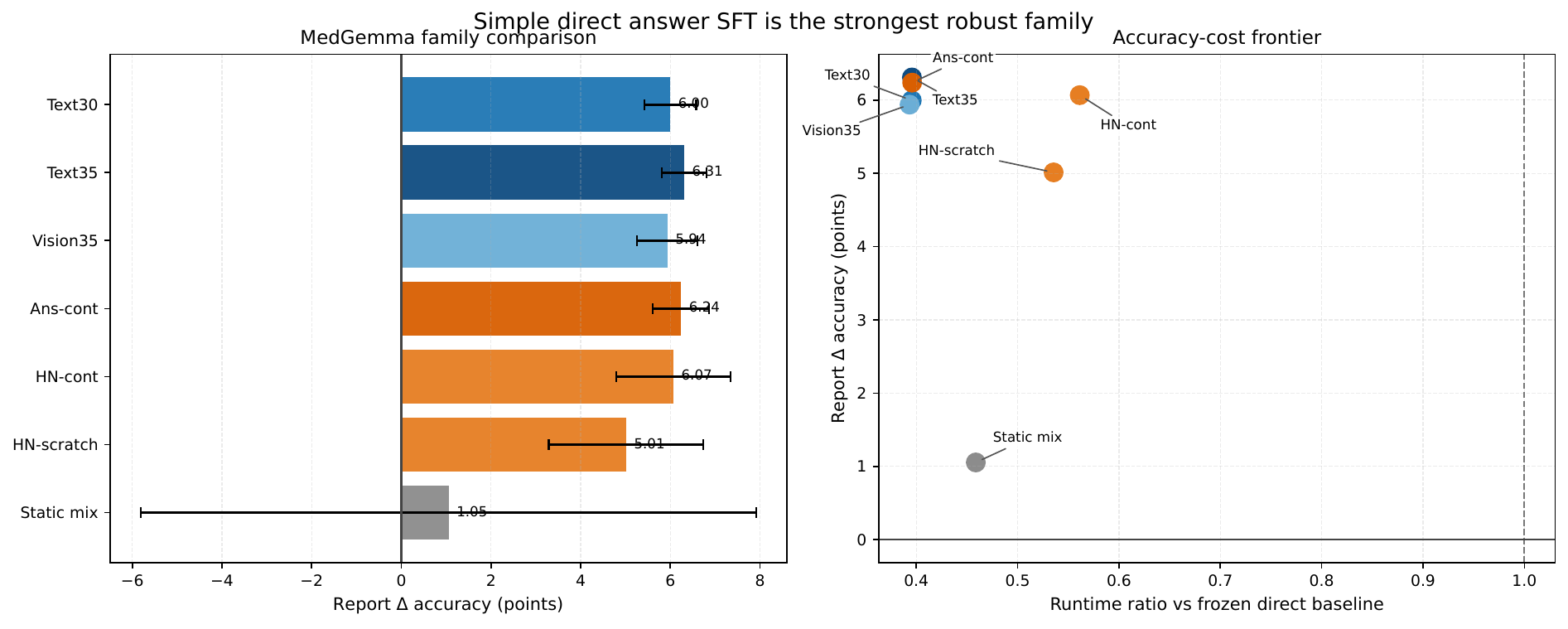}
    \caption{Family-level comparison on MedGemma. Left: report gains relative to the frozen direct baseline. Right: accuracy-cost frontier. Leader lines indicate family labels without overlapping the scatter points.}
    \label{fig:methods}
\end{figure}

\section{Experiments and Results}

We evaluate all families on the same MedFrameQA split manifest. The benchmark contains 2,851 multi-frame QA pairs \citep{yu2025medframeqa}. The primary backbone is MedGemma-1.5-4B and the secondary transfer backbone is Qwen2.5-VL-3B \citep{sellergren2026medgemma15,bai2025qwen25vl}. The held-out protocol uses \texttt{calibration\_val}=256 and \texttt{report\_test}=1024, plus fixed targeted, anchor, and hard-localization slices. The training pool contains 1,315 usable examples, with 1,187 train and 128 validation examples for the final family. The main metric is report accuracy on \texttt{report\_test}; calibration accuracy, ECE, targeted/anchor deltas, and runtime serve as secondary metrics.

The compared families are frozen direct inference, direct answer-only variants, continuation and hard-negative variants, and controller/scaffold baselines. We aggregate each family by mean and standard deviation across seeds and use paired bootstrap on the 1,024-example report split to avoid overclaiming. For the main claim, evidence is hierarchical: report accuracy and seed variance are decisive because the benchmark is scored by final answers; bootstrap among nearby direct variants determines whether the paper can justify a tuned winner or only a family winner; slice metrics, calibration, runtime, and transfer are informative but cannot override the main report criterion.

\begin{table}[t]
\centering
\caption{Main MedGemma family comparison on the 1,024-example report split. Report accuracy is shown in percent. The final tuned family is text35, but the paper's claims remain family-level because text30 and text35 are close.}
\label{tab:main-family}
\begin{adjustbox}{max width=\columnwidth}
\begin{tabular}{lrrrrrr}
\toprule
Method family & $n$ & Report Acc. & $\Delta$Acc. & Std. & $\Delta$Cal. Acc. & Runtime (s) \\
\midrule
Frozen direct baseline & 1 & 45.21 & 0.00 & -- & 0.00 & 4.14 \\
Answer-only SFT (text30) & 5 & 51.21 & 6.00 & 0.58 & 4.77 & 1.61 \\
Answer-only SFT (text35, tuned) & 5 & 51.52 & 6.31 & 0.50 & 3.20 & 1.61 \\
Answer-only + shallow vision (vision35) & 5 & 51.15 & 5.94 & 0.67 & 4.45 & 1.57 \\
Answer continuation & 8 & 51.48 & 6.24 & 0.63 & 4.05 & 1.59 \\
Hard-negative continuation & 8 & 51.31 & 6.07 & 1.27 & 2.78 & 2.30 \\
Hard-negative SFT & 6 & 50.21 & 5.01 & 1.72 & 2.28 & 2.43 \\
Static answer+hardneg mix (2:1) & 5 & 46.29 & 1.05 & 6.86 & 2.27 & 1.82 \\
\bottomrule
\end{tabular}
\end{adjustbox}
\end{table}

Guided by our robustness-aware criterion, the empirical findings across these controlled settings consistently point to a singular conclusion. The main result is that objective-aligned direct SFT is the strongest robust family on MedFrameQA. Figure~\ref{fig:methods} gives the frontier view, while Table~\ref{tab:main-family} makes the full MedGemma comparison explicit. The frozen baseline reaches 45.21\% report accuracy. The answer-only family elevates that to 51.21\% for \texttt{text30} (+6.00 points, std 0.58), 51.52\% for \texttt{text35} (+6.31, std 0.50), and 51.15\% for \texttt{vision35} (+5.94, std 0.67). Answer continuation remains competitive at 51.48\% (+6.24, std 0.63), but the harder negative families are less robust: hard-negative continuation reaches 51.31\% (+6.07) with a 1.27-point standard deviation, hard-negative SFT falls to 50.21\% (+5.01) with a 1.72-point standard deviation, and static 2:1 mixing nearly collapses to baseline at 46.29\% (+1.05) with a massive 6.86-point standard deviation. The defensible claim is therefore at the family level: direct decoder-only answer SFT is the strongest robust adaptation family we found.

That distinction between a family and a tuned point matters for the overall logic of the paper. The matched comparison shows that the direct family remains strong even under small perturbations to budget and to shallow architecture. The main finding is therefore not that one setting edges out another by a mere fraction of a point, but rather that the direct answer-only family stays on the Pareto frontier while nearby alternatives and more elaborate families fail to decisively displace it.

Paired bootstrap sharpens that interpretation. \texttt{text35} exceeds \texttt{text30} by only 0.39 points with 95\% CI $[-0.59,+1.46]$, and exceeds \texttt{vision35} by 0.29 with 95\% CI $[-0.68,+1.27]$. The evidence is strong enough to separate the direct family from clearly weaker families, but not strong enough to claim that one nearby direct variant is uniquely and decisively best.

\begin{table}[t]
\centering
\caption{Representative paired bootstrap comparisons on the 1,024-example report split. These comparisons support a family-level claim rather than a narrow tuned-variant claim.}
\label{tab:bootstrap}
\begin{adjustbox}{max width=0.8\columnwidth}
\begin{tabular}{lrrrrr}
\toprule
Comparison & $n$ & Obs. diff. & CI low & CI high & $p$ \\
\midrule
text35\_vs\_text30 & 1024 & 0.39 & -0.59 & 1.46 & 0.479 \\
text35\_vs\_vision35 & 1024 & 0.29 & -0.68 & 1.27 & 0.647 \\
text35\_vs\_answer\_old & 1024 & 0.88 & -0.49 & 2.34 & 0.241 \\
\bottomrule
\end{tabular}
\end{adjustbox}
\end{table}

More elaborate families fail to emerge as the robust winner. Answer continuation is the strongest of them, but it does not clearly beat the direct family once training budget is matched. Hard-negative continuation and hard-negative SFT illustrate the main failure mode more clearly: they preserve some slice-level usefulness, yet they pay for it with higher seed variance and slower inference. Static 2:1 answer+hard-negative mixing is the clearest instability case, with only marginal average report gain and a 6.86-point standard deviation. Controller and scaffold variants are weaker still: on a common 256-example report slice, Full-Shinka, evolved rerankers, narrow-policy control, and random-search control all remain at 46.88\% while incurring up to 4.06$\times$ runtime. These outcomes match Proposition~1 closely. Auxiliary mechanisms can improve specialist behavior, but they do not reduce global report risk enough to offset the extra bias and variance they introduce.

The slice-level behavior reinforces the same interpretation. Hard-negative continuation improves targeted and anchor slices more than answer continuation, but those gains do not translate into a stronger report family once variance is accounted for. This is the exact empirical pattern predicted by the theory: specialist improvements can coexist with worse robustness-aware model selection when the auxiliary mechanism perturbs the answer-aligned objective or increases optimization variance. This is why the paper reports a family-level result rather than a single tuned winner. A benchmark method is scientifically useful only if it remains strong under small architecture and budget perturbations, not only if it reaches one high-scoring checkpoint.

\begin{table}[t]
\centering
\caption{Held-out trade-offs across the main families. Hard-negative variants help targeted and anchor slices, but the answer-only family remains the strongest robust global method.}
\label{tab:slice-tradeoffs}
\begin{adjustbox}{max width=\columnwidth}
\begin{tabular}{lrrrrrrr}
\toprule
Family & $\Delta$Report & Std. & $\Delta$Cal. & $\Delta$Targeted & $\Delta$Anchor & $\Delta$Hard-loc & Runtime$\times$ \\
\midrule
Answer-only SFT (text30) & 6.00 & 0.58 & 4.77 & 8.33 & 1.46 & 27.34 & 0.40 \\
Answer-only SFT (text35, tuned) & 6.31 & 0.50 & 3.20 & 7.08 & 0.00 & 27.97 & 0.40 \\
Answer-only + shallow vision (vision35) & 5.94 & 0.67 & 4.45 & 7.29 & 1.46 & 26.88 & 0.39 \\
Answer continuation & 6.24 & 0.63 & 4.05 & 7.03 & -0.26 & 27.73 & 0.40 \\
Hard-negative continuation & 6.07 & 1.27 & 2.78 & 8.46 & 4.43 & 18.46 & 0.56 \\
Hard-negative SFT & 5.01 & 1.72 & 2.28 & 4.17 & 3.47 & 27.08 & 0.54 \\
Static answer+hardneg mix (2:1) & 1.05 & 6.86 & 2.27 & 4.38 & -1.87 & 29.38 & 0.46 \\
\bottomrule
\end{tabular}
\end{adjustbox}
\end{table}

Table~\ref{tab:slice-tradeoffs} makes that trade-off explicit. The direct family is not only strong on the report split; it also remains competitive on the targeted and hard-localization slices while staying substantially more stable than hard-negative variants. Hard-negative continuation is the clearest counterexample to the idea that stronger slice behavior is enough for the main claim. It is best on targeted and anchor slices, but that advantage arrives together with lower hard-localization gain, slower runtime, and more than double the seed standard deviation of \texttt{text35}. Static answer+hard-negative mixing is even more revealing: some slice numbers remain nontrivial, yet the global report metric nearly collapses and variance explodes. 

The same pattern also clarifies how to read the direct variants. \texttt{text30}, \texttt{text35}, and \texttt{vision35} all sit in a narrow band on the report frontier, and all three remain strong on hard-localization while running at roughly $0.4\times$ the frozen baseline cost. That combination is the practical signature of the main result: objective alignment does not merely win one leaderboard cell, it gives a family that remains strong under nearby budget changes, shallow architectural perturbation, and multiple evaluation slices at once.

\begin{figure}[t]
    \centering
    \includegraphics[width=\columnwidth]{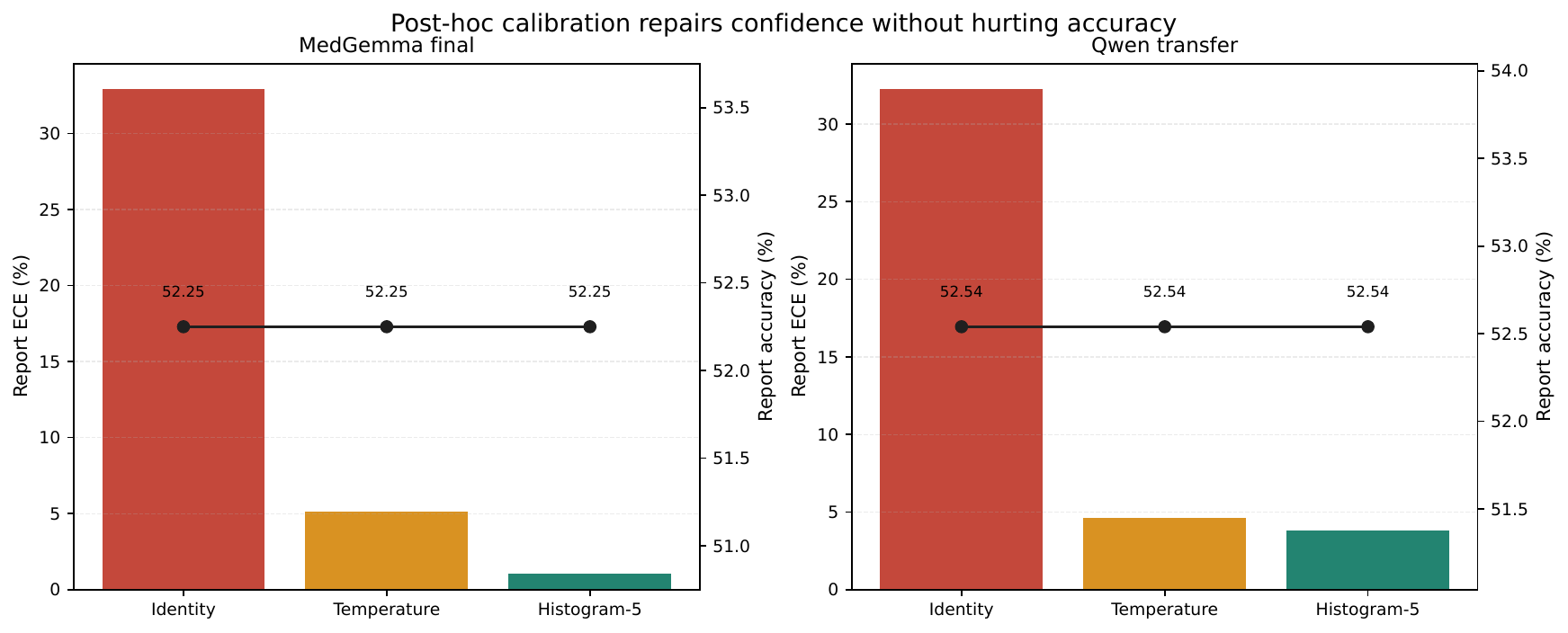}
    \caption{Post-hoc calibration of the final MedGemma model and the Qwen transfer model. Accuracy is unchanged while ECE drops sharply.}
    \label{fig:calibration}
\end{figure}

Calibration and transfer are positive but secondary. For the representative final MedGemma run, report accuracy remains 52.25\% while ECE drops from 32.93\% to 5.12\% with temperature scaling and to 1.07\% with histogram binning (Figure~\ref{fig:calibration}). This confirms that calibration is a necessary post-hoc deployment repair, not the source of the task gain. On Qwen2.5-VL-3B, the same answer-only family averages 51.99\% report accuracy, a consistent +2.58-point improvement over the frozen baseline across five runs. These results matter because they show that the family is usable after training and not wholly backbone-specific, but they do not alter the main conclusion that the direct family wins the MedFrameQA report frontier.

\section{Conclusion}

MedFrameQA appears to invite complex control, but under fixed held-out evaluation, it strongly rewards staying close to the final answer objective. Direct decoder-only answer SFT is the strongest robust family that we found: nearby direct variants remain clustered at the top of the report frontier, while continuation, hard-negative, and controller-style alternatives do not improve the main metric enough to justify their added variance and complexity. The practical recommendation is therefore simple: start from direct answer-only SFT, calibrate the probabilities post-training, and strictly demand lower held-out report risk before adopting any auxiliary reasoning machinery.

\newpage
{\small
\bibliographystyle{ieeenat_fullname}
\bibliography{references}
}

\end{document}